\documentclass{article}

\PassOptionsToPackage{numbers, compress}{natbib}

\usepackage[final]{nips_2018}

\usepackage{caption}
\usepackage{subcaption}
\usepackage{calc}
\usepackage{multicol}
\usepackage{lipsum}

\usepackage[utf8]{inputenc} 
\usepackage[T1]{fontenc}    
\usepackage{hyperref}       
\usepackage{url}            
\usepackage{booktabs}       
\usepackage{amsfonts}       
\usepackage{nicefrac}       
\usepackage{microtype}      
\usepackage{bm}
\usepackage{graphicx}
\usepackage{amsmath}
\usepackage{xr}
\usepackage{amsthm}
\usepackage[title]{appendix}
\belowdisplayskip \abovedisplayskip

\newlength{\sectionReduceTop}
\newlength{\sectionReduceBot}
\newlength{\subsectionReduceTop}
\newlength{\subsectionReduceBot}
\newlength{\abstractReduceTop}
\newlength{\abstractReduceBot}
\newlength{\captionReduceTop}
\newlength{\captionReduceBot}
\newlength{\subsubsectionReduceTop}
\newlength{\subsubsectionReduceBot}

\newlength{\eqnReduceTop}
\newlength{\eqnReduceBot}

\newlength{\horSkip}
\newlength{\verSkip}

\newlength{\figureHeight}
\title{Conditional Computation for Continual Learning}

\author{
  Min Lin \\
  Mila, Université de Montréal \\
  \And
  Jie Fu \\
  Mila, Université de Montréal \\
  IVADO \\
  \AND
  Yoshua Bengio \\
  Mila, Université de Montréal \\
  CIFAR Senior Fellow \\
  \texttt{} \\
}

\begin{document}

\maketitle

\vspace{\abstractReduceTop}
\begin{abstract}
\vspace{\abstractReduceBot}
  Catastrophic forgetting of connectionist neural networks is caused by the global sharing of parameters among all training examples. In this study, we analyze parameter sharing under the conditional computation framework where the parameters of a neural network are conditioned on each input example. At one extreme, if each input example uses a disjoint set of parameters, there is no sharing of parameters thus no catastrophic forgetting. At the other extreme, if the parameters are the same for every example, it reduces to the conventional neural network. We then introduce a clipped version of \textit{maxout} networks which lies in the middle, i.e. parameters are shared partially among examples. Based on the parameter sharing analysis, we can locate a limited set of examples that are interfered when learning a new example. We propose to perform rehearsal on this set to prevent forgetting, which is termed as \textit{conditional rehearsal}. Finally, we demonstrate the effectiveness of the proposed method in an online non-stationary setup, where updates are made after each new example and the distribution of the received example shifts over time.
\end{abstract}

\vspace{-2pt}
\vspace{\sectionReduceTop}
\section{Introduction}
\vspace{\sectionReduceBot}

In this paper, we study the level of parameter sharing within the conditional computation framework~\cite{Bengio2013,Cho2014,Bengio2015,Shazeer2017}. The key idea of conditional computation is to make the parameters $\theta$ of the neural network $f$ a function of the input $x$ denoted as $\Theta(x)$. The computation defined by $f^\theta$ is conditioned on $x$ through the function $\Theta(x)$:
\vspace{-2pt}
\begin{equation}
  y = f^\theta(x) = f^{\Theta(x)}(x)
\end{equation}

The conventional neural network has $\Theta(x)=\theta^C$. The network parameter $\theta^C$ is independent of $x$, and therefore it is shared globally by all examples. To reduce the level of parameter sharing thus reducing forgetting, we need to choose $\Theta$ other than constant functions.

A na\"ive choice of $\Theta$ to prevent forgetting is a look-up table $T_{x\rightarrow\theta}$ that keeps different parameters for every unique input $x$:
\vspace{-2pt}
\begin{equation}
  \label{eqn:non_param}
  \Theta(x) = T_{x\rightarrow\theta}[x]
\end{equation}

With this one-to-one mapping between $x$ and $\theta$, there is zero parameter sharing between examples, and thus no forgetting will happen when learning new examples. However, this choice is not very interesting as it results in a local non-parametric $f^\theta$, losing the generalization property of neural networks.

Given neural networks and local non-parametric models as two extremes in the conditional computation framework, we seek to strike a balance between those two extremes in the hope that the resulting model can take the best of both worlds, i.e. having good generalization like neural networks while suffering less from catastrophic forgetting like local non-parametric models.

The contributions of this work are:
\begin{itemize}
  \item We analyze parameter sharing and correspondingly the interfered examples when learning new knowledge under the conditional computation framework.
  \item Based on the analysis of interfered example, we propose conditional rehearsal to rehearse only the interfered examples, which is more efficient than random rehearsal~\cite{Robins1995}.
  \item We introduce clipped maxout, which has a smaller set of interfered examples when learning a new example, compared to maxout. Together with conditional rehearsal, it is capable of continual learning.
  \item We also evaluate our proposed method in a new setup of MNIST, named \textit{MNIST-ol}, where a single example is used for training at a time, and the distribution of the received example shifts over time.
\end{itemize}

\vspace{-2pt}
\vspace{\sectionReduceTop}
\section{Conditional Computation with Partial Parameter Sharing}

\subsection{Many-to-One Mapping between $x$ and $\theta$}
\label{subsection:many_to_one}
Consider first how one could share parameters within groups of examples using
\vspace{-2pt}
\begin{equation}
  \label{eqn:many_to_one}
  \Theta(x) = T_{G(x)\rightarrow{\theta}}[G(x)]
\end{equation}

We would map the examples to a group id through the grouping function $G(x)$ and then associate unique parameters to each group through a look-up table. In contrast to the one-to-one mapping defined in Eqn. \ref{eqn:non_param}, in Eqn. \ref{eqn:many_to_one} we define a many-to-one mapping between examples and parameters. Parameters are shared between examples mapping to the same group. To reduce complication, we assume that the grouping function $G$ is pre-defined so that the parameter sharing relationships between examples are fixed. The situation where $G$ changes will be revisited in Sec. \ref{sec:maxout}. Under this setting, learning of a new example interferes only with historical examples from the same group of the new example.

Additionally, it is also computationally more efficient to do rehearsal since we only need to rehearse over a limited number of examples that are interfered, i.e. those in the same group of the new example. We term this as conditional rehearsal because the rehearsal set is conditioned on the new example being learned. Note that in this work we assume all historical examples are available, and the examples can be stored in a look-up table indexed by $G(x)$ for fast retrieval.



\subsection{Many-to-Many Mapping between $x$ and $\theta$}
The many-to-one mapping assumes that examples belonging to different groups cannot share parameters, which may be too restrictive and loses the combinatorial advantage of sharing enjoyed by deep neural
networks~\citep{montufar2014number}. We can easily extend Eqn. \ref{eqn:many_to_one} to a many-to-many mapping by assigning more than one group id to the input $x$:
\vspace{-2pt}
\begin{equation}
\label{eqn:many_to_many}
  \Theta(x)=\{\cdots, T_{G_i(x)\rightarrow\theta}[G_i(x)], \cdots \} 
\end{equation}


\subsubsection{Maxout Network as Conditional Computation}
\label{sec:maxout}
One empirical argument is that we can reduce forgetting if we sparsify the update to the weights by means of \textit{node sharpening}~\cite{French1991}, \textit{dropout}~\cite{Goodfellow2013}, maxout~\cite{goodfellow2013maxout} or \textit{compete to compute}~\cite{Srivastava2013}. Although empirically they do exhibit a slower forgetting property, the results are still far from satisfactory~\cite{Goodfellow2013}. Here we analyze the forgetting property of maxout networks under the conditional computation framework and introduce a few modifications to further reduce forgetting.

A maxout unit implements the following function:
\vspace{-2pt}
\begin{equation}
    h_i(x)=\max_{j\in{[1,k]}}x^TW[:,i,j]+b[i,j]
\end{equation}

where $W\in{R^{d\times m\times k}}$ and $b\in{R^{m\times k}}$ are the parameters so that there are $m$ outputs and each output is the maximum over $k$ neurons. It can be transformed into the conditional computation form:
\vspace{-2pt}
\begin{equation}
    \label{eqn:maxout_Theta}
    \Theta(x)=\left\{W[:,i,G_i(x)], b[i,G_i(x)]\mid i\in{[1,m]} \right\}
\end{equation}
\vspace{-2pt}
\begin{equation}
    \label{eqn:maxout_G}
    G_i(x)=arg\max_j x^TW[:,i,j]+b[i,j]
\end{equation}

\begin{figure}
\centering
\begin{minipage}[t]{0.48\textwidth}
  \centering
  \vspace{0pt}
  \includegraphics[width=1.0\linewidth]{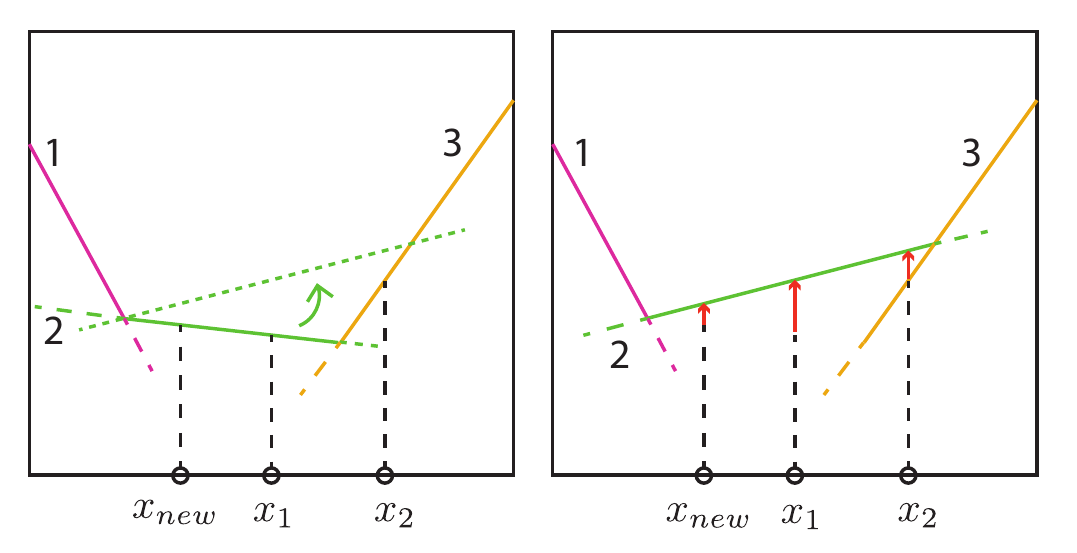}
  \captionof{figure}{1D demonstration of the change when parameter gets updated. The left figure shows $G(x_{new})=G(x_1)=2$ and $G(x_2)=3$. Learning of $x_{new}$ pushes line $2$ up. The right figure shows that the update interferes not only with $x_1$ but also $x_2$, i.e. $G(x_2)=2$ after the update. }
  \label{fig:maxout_massive_interfere}
\end{minipage}%
\hspace{0.3cm}
\begin{minipage}[t]{0.48\textwidth}
  \centering
  \vspace{0pt}
  \includegraphics[width=.8\linewidth]{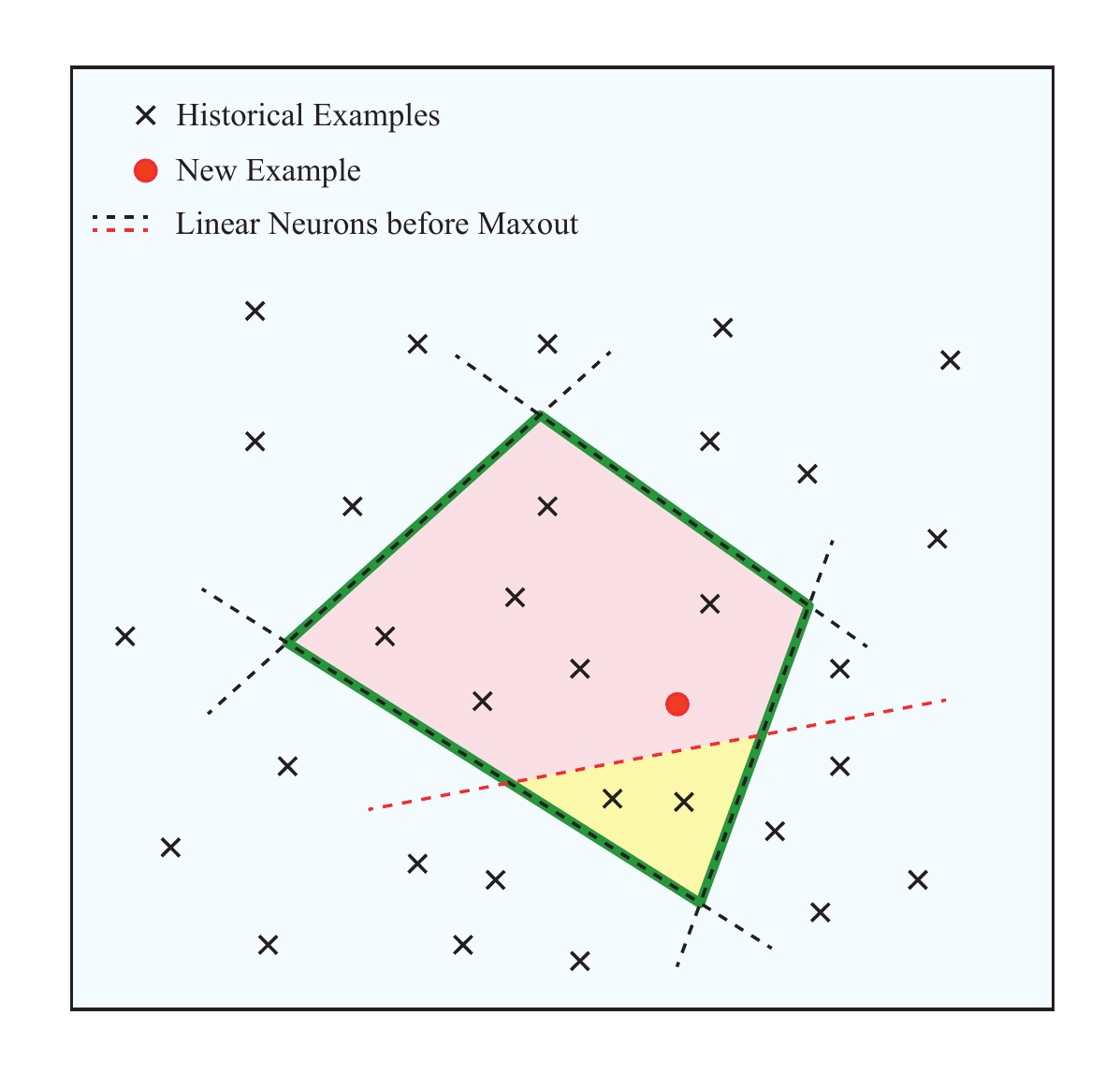}
  \captionof{figure}{2D schematic about how learning on a new example interferes with historical examples for clipped maxout.}
  \label{fig:maxout_with_clip}
\end{minipage}
\end{figure}

We use $\mathcal{S}$ to represent the set of all historical examples, and $\hat{\mathcal{S}}$ for the interfered examples. Take Eqn. \ref{eqn:maxout_Theta} alone, if $G_i$ is pre-defined and fixed, learning $x_{new}$ only interferes with examples in $\left\{x_{old}\mid x_{old}\in{\mathcal{S}}; \forall i\in[1,m], G_i(x_{old})=G_i(x_{new})\right\}$ denoted by $\hat{\mathcal{S}}_\text{fix-G}$. However, the assumption that $G_i$ is fixed does not hold because $G_i$ itself uses $W$ and $b$ as parameters. When $W$ and $b$ are updated, $G_i(x)$ and thus $h_i(x)$ is potentially massively modified for any $x$, including when $x\notin{\hat{\mathcal{S}}_\text{fix-G}}$. This can be demonstrated with the 1D case in Fig. \ref{fig:maxout_massive_interfere}.
For simplicity, we omit the index $i$ here and for the rest of the paper when discussing only one maxout unit with scalar output. In this figure, we show that although $G(x_2)\neq 2$, its output value is still changed when linear neuron $2$ gets updated. Note that if the change to linear neuron $2$ is small enough, there is probability that $G(x_2)$ does not change. This could be the reason why empirically maxout slightly mitigates the forgetting as shown in~\cite{Goodfellow2013}, but there is no theoretical guarantee.

Therefore, in the worst case of linear maxout, the interfered set $\hat{\mathcal{S}}$ equals $\mathcal{S}$. To make $\hat{\mathcal{S}}$ strictly smaller than $\mathcal{S}$, we introduce a few modifications to maxout in the next section.

\subsubsection{Clipped Maxout and Conditional Rehearsal}

We clip the linear output before maxout to a constant $C$ with $\min(\cdot,C)$. $C$ can be any fixed value because we can rely on the bias term $b$ to control the magnitude of the output. We use $\min(\cdot, 0)$ for clipping in this paper. The clipped maxout is described with the following function:
\vspace{-2pt}
\begin{align}
h(x) &= \max_{j\in{[1,k]}}z_j(x) \\
z_j(x) &= \min(x^TW[:,j]+b[j], 0)
\end{align}

Conditioned on the new example $x_{new}$, the historical examples set $\mathcal{S}$ can be divided into 3 disjoint sets based on the value of $z_j(x)$:
\begin{enumerate}
    \item $\mathcal{S}_1=\{x \mid \forall j, z_j(x)<0; x\in{\mathcal{S}}\}$.
    \item $\mathcal{S}_2=\{x \mid \forall j\neq{G(x_{new})}, z_j(x)<0; z_{G(x_{new})}(x)=0; x\in{\mathcal{S}}\}$
    \item $\mathcal{S}_3=\{x \mid \exists j\neq{G(x_{new})}, z_j(x)=0; x\in {\mathcal{S}} \}$
\end{enumerate}
We show a 2D graphical depiction of the 3 sets in Fig. \ref{fig:maxout_with_clip}. $\mathcal{S}_1$ is in pink, $\mathcal{S}_2$ in yellow and $\mathcal{S}_3$ in light blue. The dashed red line stands for the linear neuron selected by $G(x_{new})$, it will be updated when we perform one step of gradient descent on $x_{new}$. When the update happens, only examples falling in $\mathcal{S}_1$ and $\mathcal{S}_2$ will be interfered. Examples in $\mathcal{S}_3$ will not be interfered because they are clipped on at least one neuron that are not updated. Refer to Sec. \ref{sec:app_proof} of the Appendix for a formal proof. One good property of the clipped maxout is that the interfered examples falls within the convex set enclosed by the linear neurons excluding the neuron being updated. This convex set could potentially be small if enough neurons are maxed out.

Given that training on the new example only interferes with $\hat{\mathcal{S}}=\mathcal{S}_1\cup\mathcal{S}_2$, we can utilize conditional rehearsal to specifically rehearse these examples when learning new knowledge. If the model has enough capacity to learn new knowledge and at the same time preserve the output for the rehearsed examples, it is guaranteed that there will be no forgetting of historical examples. The effectiveness of conditional rehearsal on clipped maxout units will be verified in the experiments section.

\vspace{-2pt}
\subsubsection{Minimally Clipped Minout}
To make the activation value positive rather than negative in the convex set $\mathcal{S}_1$, we adopt the mirror negative of the maximally clipped maxout, which is the minimally clipped \textit{minout}. The definition and the motivation of minimally clipped minout is detailed in Sec. \ref{sec:app_minout} of the Appendix.

\vspace{-2pt}
\vspace{\sectionReduceTop}
\section{Experiments}
\vspace{\sectionReduceBot}

\textbf{Data ---} We experiment on the MNIST dataset in this work.

\textbf{Setups ---} \textit{Disjoint MNIST}~\cite{Srivastava2013} and \textit{Permuted MNIST}~\cite{Goodfellow2013,Kirkpatrick2017} are the most commonly used settings. Disjoint MNIST splits the dataset into multiple subsets which have disjoint labels. Permuted MNIST creates new datasets from MNIST by permuting the pixels. For these two setups, the algorithm is trained on one subset at a time with i.i.d. assumption within each subset.

In this work, however, we study continual learning with an online non-stationary setting where a \emph{single} example at a time is seen before making an update, and the distribution of the received example shifts over time. The goal is to fit optimally to the already seen examples at any time point of the training, which can be measured by the accuracy on the test set throughout the training procedure. Accordingly, we propose a new setup for MNIST dataset named as MNIST with ordered labels (MNIST-ol). As the name implies, the training images are arranged by their associated labels in an ascending (or descending) order. For example, during training, images with label $0$ are received first, and those with label $9$ are received last. Ordering by labels removes the assumption that each example from the data stream is drawn i.i.d. from the whole training set. It can be seen from Fig. \ref{fig:mnist_ol_vs_iid} in the Appendix that learning with stochastic gradient descent completely fails on MNIST-ol.

Due to current suboptimal implementation, we experiment with a subset of MNIST-ol by randomly taking 100 examples from each class.

\textbf{Model configuration ---} Our model is a single layer minout network with 10 minout units each corresponding to one label. Each minout unit has 50 linear neurons. We apply \textit{sigmoid} activation function on the linear neurons before minout so that they are clipped to $[0, 1]$. The output of each minout unit is directly used as the probability of each label and trained by a per label \textit{sigmoid} cross entropy loss. Activation value smaller than $0.1$ is seen as clipped to $0$ when deciding the interfered set.

\textbf{Baselines ---} For baseline, we compare to the same model trained on MNIST-ol without rehearsal and with rehearsal on randomly selected historical examples. The number of randomly selected examples are set to match the number conditionally rehearsed examples in the studied method, which is 100 according to Sec. \ref{sec:num_rehearsed}.

\textbf{Training ---} Training happens one example after another with an additional rehearsal loss and corresponding updates. For both this method and the baselines, the training of an example on one maxout unit is stopped as soon as the loss on this unit is smaller than $0.1$.

\subsection{Number of Examples Rehearsed}
\label{sec:num_rehearsed}
\begin{figure}
    \centering
    \includegraphics[width=0.45\textwidth]{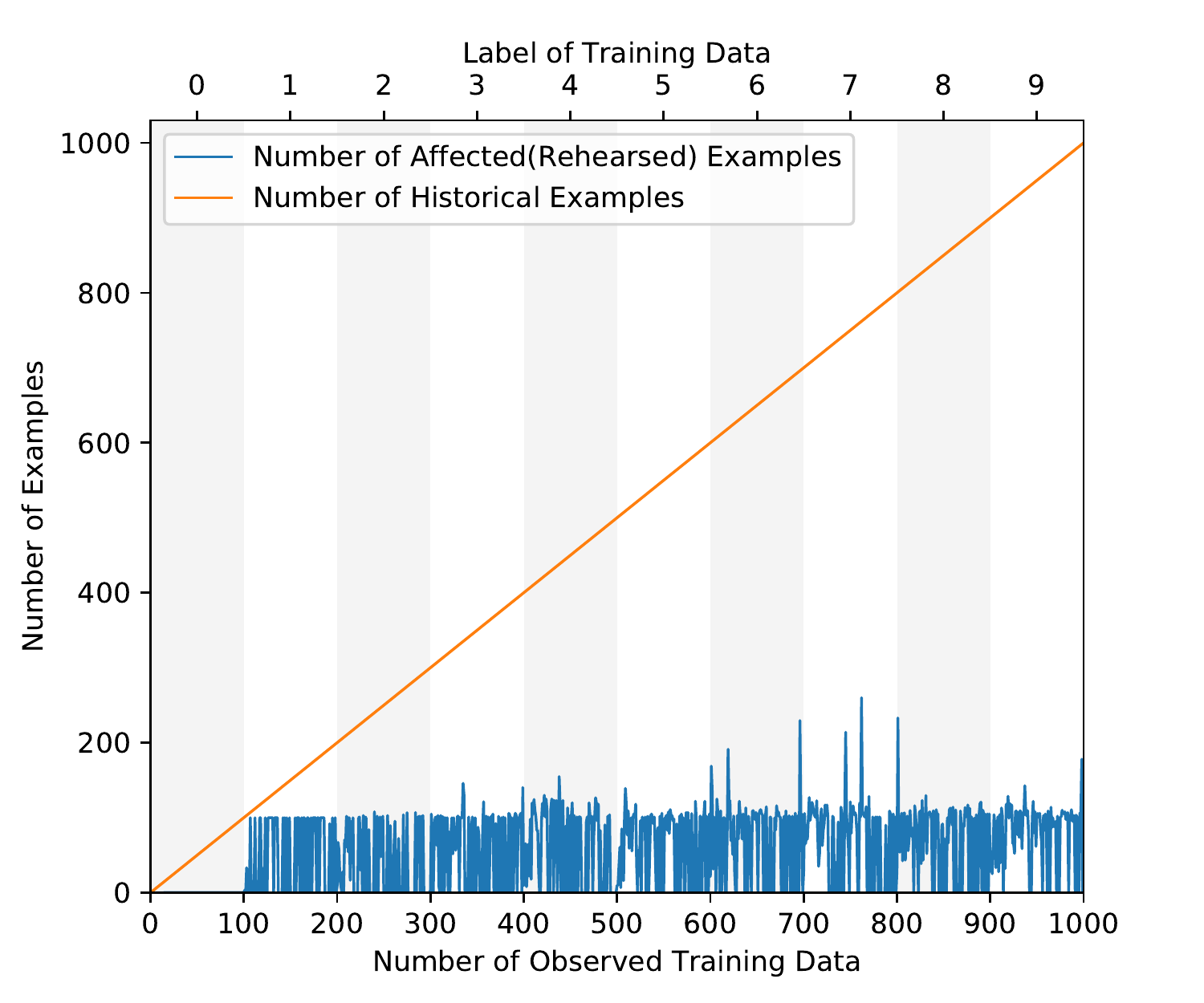}
    \caption{Number of rehearsed examples throughout training.}
    \label{fig:num_rehearsed_examples}
\end{figure}

Under the configuration of our model, each minout unit encloses one class of the training data in the convex set $\mathcal{S}_1$, which suggests that the theoretical number of rehearsed examples should be around 100. To verify this, we plot the average number of examples that are rehearsed for each minout unit during training in Fig. \ref{fig:num_rehearsed_examples}. It can be seen that the number of rehearsed data throughout training fluctuates around 100, which is consistent with our expectations. More discussion on the number of rehearsed examples can be found in Appendix \ref{sec:app_num_rehearse}

\subsection{Accuracy and Forgetting Behavior of the Proposed Method}
\begin{figure}
\centering
\begin{minipage}[t]{0.45\textwidth}
  \centering
  \vspace{0pt}
  \includegraphics[width=1.0\linewidth]{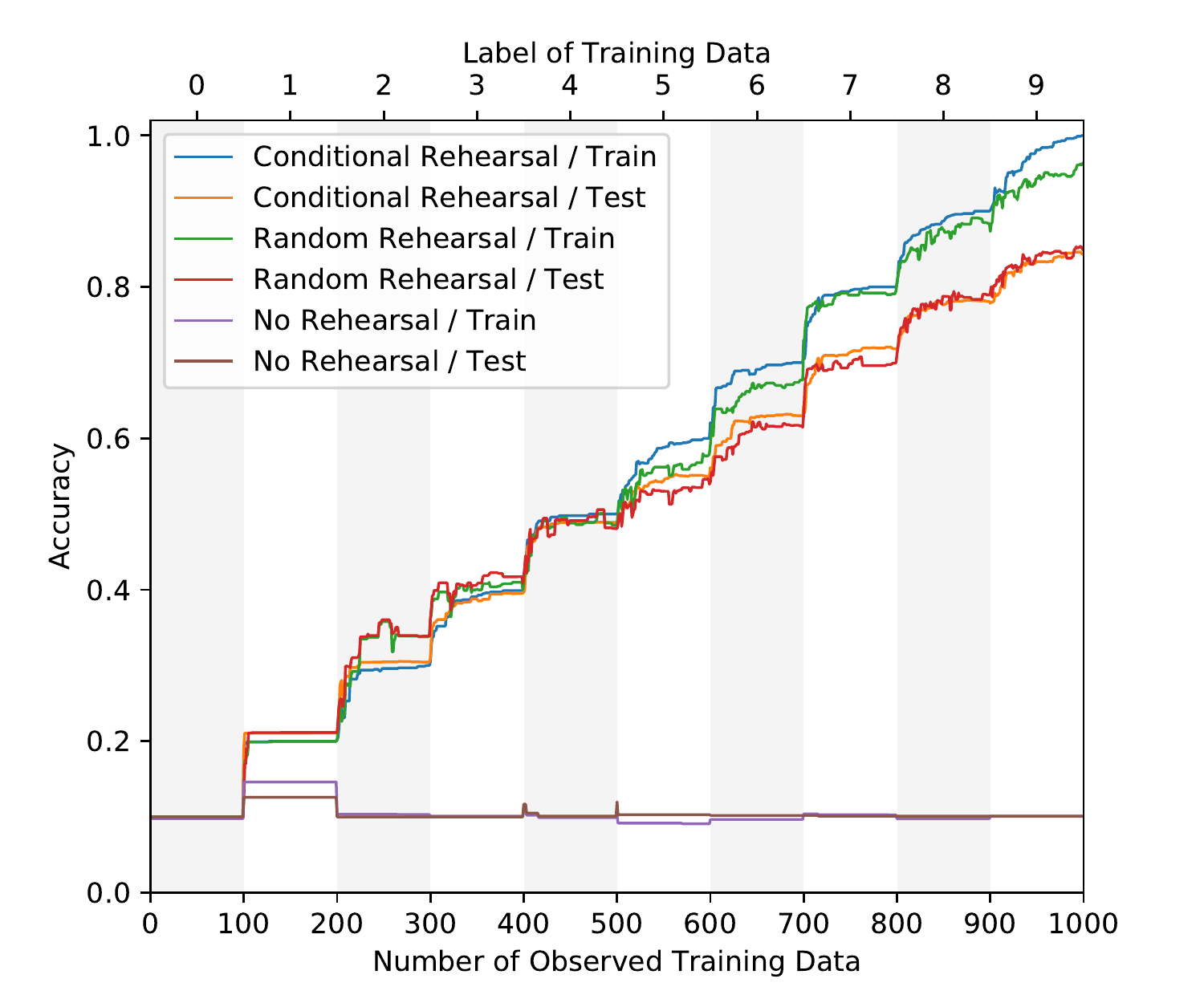}
  \captionof{figure}{Accuracy of conditional rehearsal \textit{vs} random rehearsal \textit{vs} no rehearsal.}
  \label{fig:acc}
\end{minipage}%
\hspace{0.3cm}
\begin{minipage}[t]{0.45\textwidth}
  \centering
  \vspace{0pt}
  \includegraphics[width=1.0\linewidth]{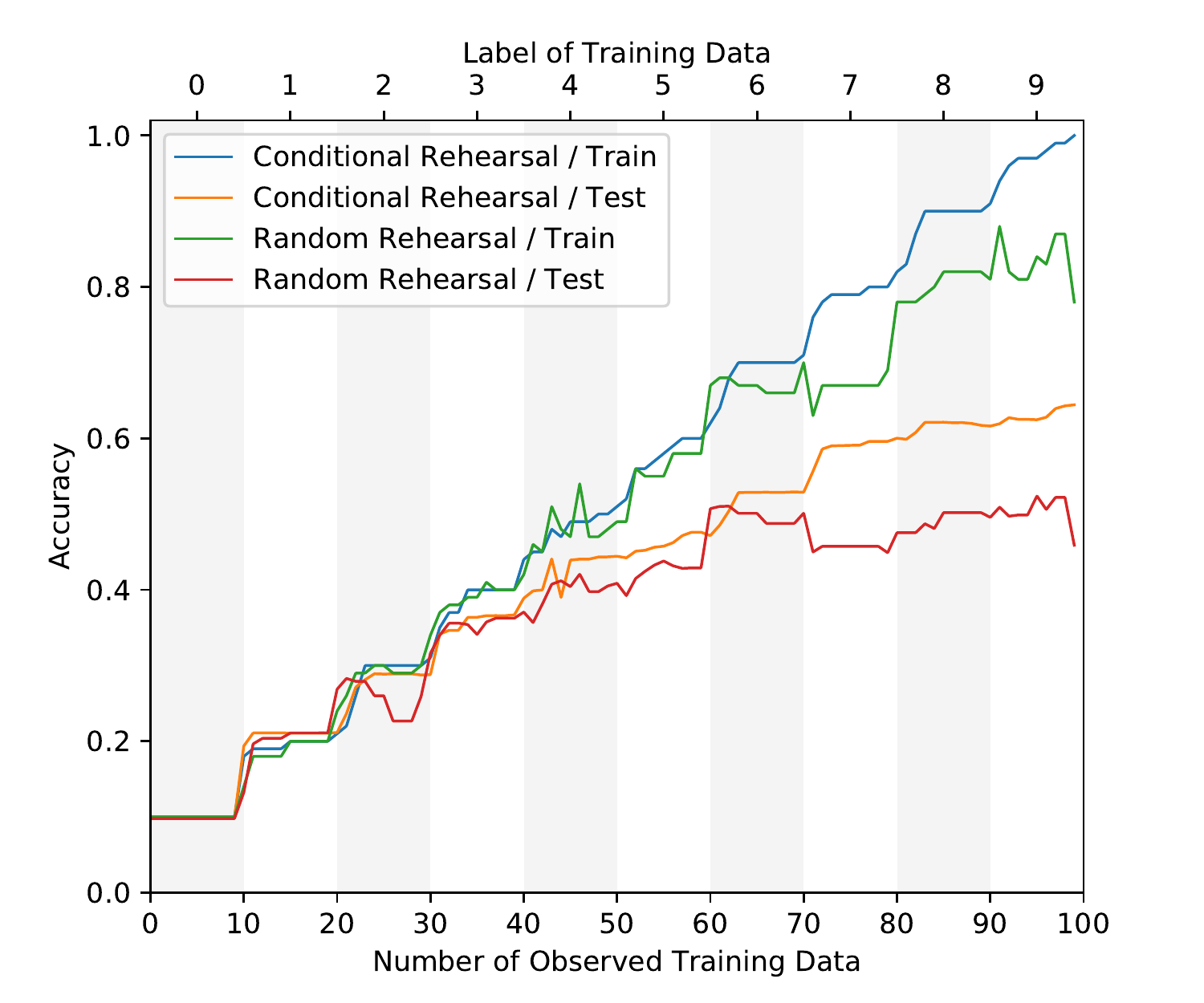}
  \captionof{figure}{Accuracy of conditional rehearsal \textit{vs} random rehearsal with a smaller training set.}
  \label{fig:acc_smaller_scale}
\end{minipage}
\end{figure}

We test the accuracy of both the training set and test set after learning of every example, and plot the training/testing accuracy in Fig. \ref{fig:acc}. We can see that both training and testing accuracy monotonically increase throughout training for clipped minout with conditional rehearsal. Accuracy on the training set reached 100\% at the end of training, which means no forgetting is happening.

The no rehearsal baseline fails to learn as expected. However, it seems that the random rehearsal baseline is doing as good as the conditional rehearsal. We argue that this is because the MNIST dataset has only a few modes and that 100 randomly selected examples would contain enough information of the whole dataset. For a more complex dataset where the number of modes exceeds the number of rehearsed examples, conditional rehearsal would be advantageous because it is more selective and thus more efficient. We verify this by further reducing the training set to 10 examples from each class. Correspondingly the number of rehearsal examples is reduced to 10. It is harder for 10 examples to contain sufficient information of the whole dataset. As is shown in Fig. \ref{fig:acc_smaller_scale}, conditional rehearsal outperforms random rehearsal by a big margin.

\vspace{-2pt}
\vspace{\sectionReduceTop}
\section{Future Directions}
\vspace{\sectionReduceBot}

We will focus on two directions in the future. First, we aim to develop a \textit{deep} version of the proposed clipped maxout network. Second, we plan to design connectionist approaches for storing historical data.

\bibliographystyle{unsrt}
\bibliography{nips_2018}

\begin{appendices}









\section{One-step update on $x_{new}$ does not interfere with examples in $\mathcal{S}_3$}

Denote the parameter after updating on $x_{new}$ as $W'=W+\Delta{W}$ and $b'=b+\Delta{b}$, where $\Delta{W[:,j]}=0$ and $\Delta{b[:,j]}=0$ if $j\neq{}G(x_{new})$.

\newtheorem{thm}{Theorem}
\begin{thm}
Let $h'(x)$ denote the function $h(x)$ after the update, then $h'(x)=h(x)$ for $\forall{x\in\mathcal{S}_3}$.
\end{thm}
\textit{Proof.} By definition of $\mathcal{S}_3$, for $x\in{\mathcal{S}_3}$ there exists $i\neq{G(x_{new})}$ where $z_i(x)=0$. 

For any $j$, $$z_j(x)=\min(x^TW[:,j]+b[j], 0)\leq{}z_i(x)$$
Therefore, $h(x)=\max_{j\in{[1,k]}}{z_j(x)}=z_i(x)=0$. 

Let $z'_i(x)$ denote the $z_i(x)$ after the update, 
\begin{align*}
z'_i(x) &= \min(x^TW'[:,i]+b'[i], 0) \\
&= \min(x^T(W[:,i]+\Delta{W}[:,i])+b[i]+\Delta{b}[i], 0) \\
&= \min(x^TW[:,i]+b[i], 0) \\
&= z_i(x) \\
&= 0
\end{align*}

Therefore, $h'(x)=\max_{j\in{[1,k]}}{z'_j(x)}=z'_i(x)=0=h(x)$, which completes the proof.

\section{Bookkeeping for Conditional Rehearsal}
To efficiently locate the interfered historical examples when training a new example, we use a key-value store to keep the historical data. The keys are indice of linear neurons, and an example is stored under a key if the corresponding neuron is clipped at this example. At the same time a counter is associated with each example to count how many linear neurons are clipped. For a historical example, if no neurons are clipped, it falls in $\mathcal{S}_1$; if only the neuron selected by the new example is clipped, it falls in $\mathcal{S}_2$; otherwise it falls in $\mathcal{S}_3$. After the weight update, the information in the table is updated accordingly.

\section{Minimally Clipped Minout}
The minimally clipped minout unit is defined as follows,

\begin{align}
h(x) &= \min_{j\in{[1,k]}}z_j(x) \\
z_j(x) &= \max(x^TW[:,j]+b[j], 0)
\end{align}

We can rewrite Eqn. 8 and Eqn. 9 as the negative of minout, with $-W$ and $-b$ as the parameters.

\begin{align}
h(x) &= \max_{j\in{[1,k]}}z_j(x) = -\min_{j\in{[1,k]}}-z_j(x) \\
-z_j(x) &= -\min(x^TW[:,j]+b[j], 0) = \max(-x^TW[:,j]-b[j], 0)
\end{align}

In this work we adopt minimally clipped minout instead of maximally clipped maxout because it is activated (larger than 0) rather than deactivated (smaller than 0) in the convex set $\mathcal{S}_1$. This aligns better with human instinct. When we think of the maxout unit as a detector of some property of the input data, we would like the unit be activated when the property is present.

\section{Relationship between Minout and the Gating mechanism for Conditional Computation}
The original conditional computation paper and follow-up works introduce binary gating neurons to turn on/off computing neurons conditioned on inputs \cite{Serra2018}. In practice, \textit{sigmoid} activation functions are usually used in place of the binarization for the ease of training. Assuming both the computing neuron and gating neuron use \textit{sigmoid} activation functions, it can be written as:

\begin{equation}
\label{eqn:gating}
y=\underbrace{\sigma(x^TW_1+b_1)}_\text{Computing neuron}\odot{}\underbrace{\sigma(x^TW_2+b_2)}_\text{Gating neuron}
\label{eqn:sup:gating}
\end{equation}
Note that the computing neuron and gating neuron are indistinguishable and can be swapped. We can generalize this into $y=\prod_j{\sigma(x^TW_j+b_j)}$. We can see that $y=0$ if there exists $j$ so that $\sigma(x^TW_j+b_j)=0$. And $y>0$ only when $\sigma(x^TW_j+b_j)>0$ for $any$ $j$.

For clipped Minout:

\begin{equation}
   y=\min_{j\in{[1,k]}}\max(x^T W_{j}+b_{j}, 0)
   \label{eqn:sup:clipped_minout}
\end{equation}
It has a similar behavior, $y>0$ only when \textit{all} of $\max(x^T W_{j}+b_{j}, 0)>0$.

In fact, Eqn. \ref{eqn:sup:gating} and \ref{eqn:sup:clipped_minout} can be seen as AND functions, whose output is non-zero only when all the neurons are activated.

\section{Number of Rehearsed Examples}
One can consider the maxout unit as a detector of some property of the input. Inputs that have this property will fall within the convex set $\mathcal{S}_1$, we call them positive examples. Inputs that do not have this property will fall outside of the convex set, and we call them negative examples.

For a single maxout unit, the number of the rehearsed examples depends on the size of $\mathcal{S}_1$ and $\mathcal{S}_2$. $\mathcal{S}_2$ can be small if most negative examples are clipped at more than one neuron. The size of $\mathcal{S}_1$ depends on how many historical examples activates this maxout unit. In this paper, since each maxout unit corresponds directly to one of the categories, the size of $\mathcal{S}_1$ is approximately one tenth (for ten categories in MNIST) of the total training examples. 

In the future, we can have more maxout units in the hidden layers in the future with a deep version of this idea. And we can study the relationship between the number of rehearsed examples and the number of maxout units.

\section{Stochastic Gradient Descent fails on MNIST-\textit{ol}}

A 2-layer multilayer perceptron ($784\rightarrow128\rightarrow10$) with ReLU activations and a softmax loss is constructed and trained with SGD on MNIST-ol.

\begin{figure}[h]
    \centering
    \includegraphics[width=0.8\textwidth]{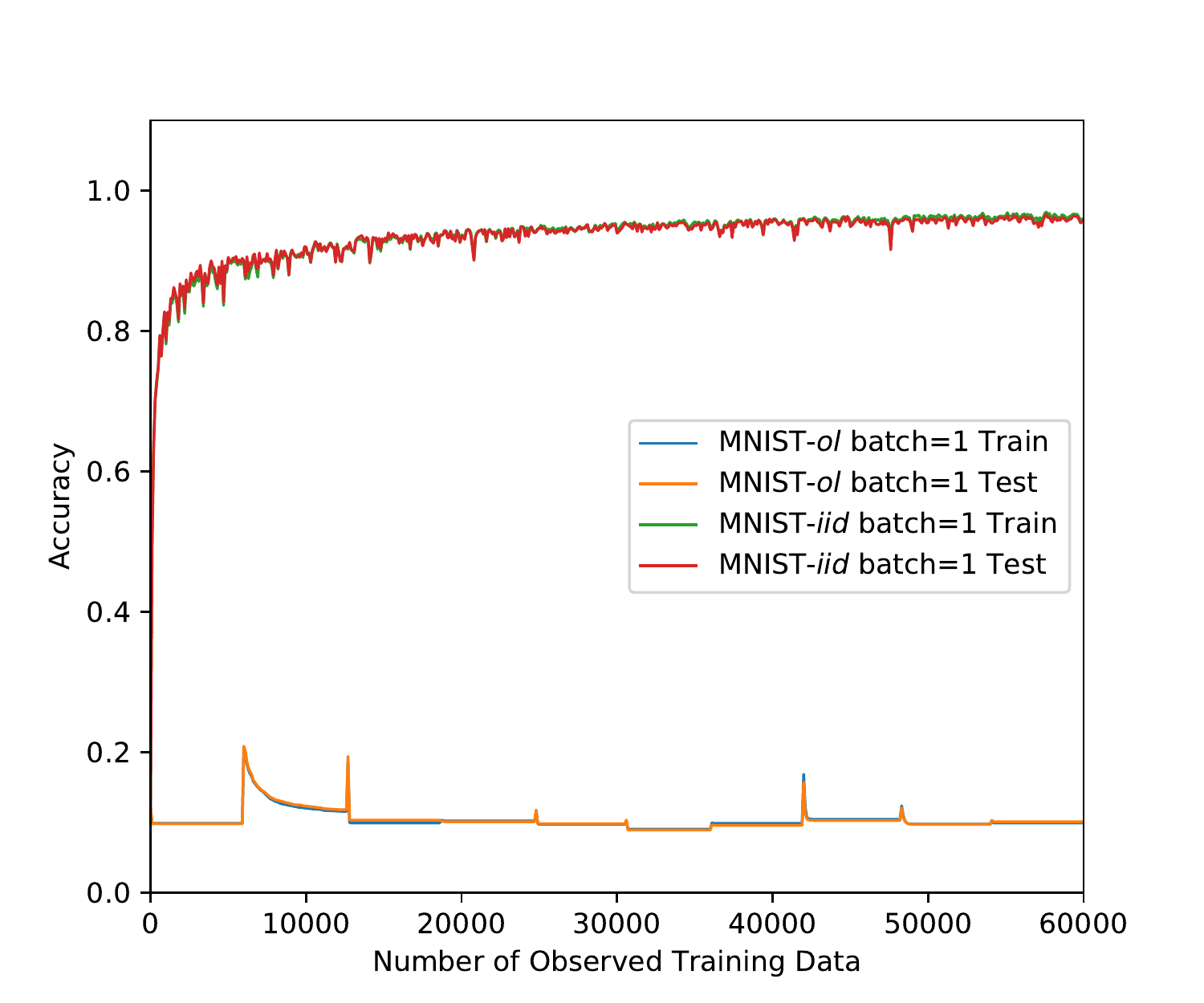}
    \caption{Training with Stochastic Gradient Descent fails on MNIST-ol.}
    \label{fig:mnist_ol_vs_iid}
\end{figure}

Fig. \ref{fig:mnist_ol_vs_iid} shows that na\"ively training with SGD fails utterly on MNIST-ol task, whereas our proposed method does not suffer from catastrophic forgetting. 



\section{One-step update on $x_{new}$ does not interfere with examples in $\mathcal{S}_3$}
\label{sec:app_proof}
Denote the parameter after updating on $x_{new}$ as $W'=W+\Delta{W}$ and $b'=b+\Delta{b}$, where $\Delta{W[:,j]}=0$ and $\Delta{b[:,j]}=0$ if $j\neq{}G(x_{new})$.

\newtheorem{thm}{Theorem}
\begin{thm}
Let $h'$ denote the function $h$ after the update, then $h'(x)=h(x)$ for $\forall{x\in\mathcal{S}_3}$.
\end{thm}
\textit{Proof.} By definition of $\mathcal{S}_3$, for $x\in{\mathcal{S}_3}$ there exists $i\neq{G(x_{new})}$ where $z_i(x)=0$.

For any $j$, $$z_j(x)=\min(x^TW[:,j]+b[j], 0)\leq{}z_i(x)$$
Therefore, $h(x)=\max_{j\in{[1,k]}}{z_j(x)}=z_i(x)=0$.

Let $z'$ denote the $z$ after the update,
\begin{align*}
z'_i(x) &= \min(x^TW'[:,i]+b'[i], 0) \\
&= \min(x^T(W[:,i]+\Delta{W}[:,i])+b[i]+\Delta{b}[i], 0) \\
&= \min(x^TW[:,i]+b[i], 0) \\
&= z_i(x) \\
&= 0
\end{align*}

Therefore, $h'(x)=\max_{j\in{[1,k]}}{z'_j(x)}=z'_i(x)=0=h(x)$, which completes the proof.

\section{Bookkeeping for Conditional Rehearsal}
\label{sec:app_bookkeep}
To efficiently locate the interfered historical examples when training a new example, we use a key-value store to keep the historical data. The keys are indice of linear neurons, and an example is stored under a key if the corresponding neuron is clipped at this example. At the same time a counter is associated with each example to count how many linear neurons are clipped. For a historical example, if no neurons are clipped, it falls in $\mathcal{S}_1$; if only the neuron selected by the new example is clipped, it falls in $\mathcal{S}_2$; otherwise it falls in $\mathcal{S}_3$. After the weight update, the information in the table is updated accordingly.

\section{Minimally Clipped Minout}
\label{sec:app_minout}
The minimally clipped minout unit is defined as follows,

\begin{align}
h(x) &= \min_{j\in{[1,k]}}z_j(x) \\
z_j(x) &= \max(x^TW[:,j]+b[j], 0)
\end{align}

We can rewrite Eqn. 8 and Eqn. 9 as the negative of minout, with $-W$ and $-b$ as the parameters.

\begin{align}
h(x) &= \max_{j\in{[1,k]}}z_j(x) = -\min_{j\in{[1,k]}}-z_j(x) \\
-z_j(x) &= -\min(x^TW[:,j]+b[j], 0) = \max(-x^TW[:,j]-b[j], 0)
\end{align}

In this work we adopt minimally clipped minout instead of maximally clipped maxout because it is activated (larger than 0) rather than deactivated (smaller than 0) in the convex set $\mathcal{S}_1$. This aligns better with human instinct. When we think of the maxout unit as a detector of some property of the input data, we would like the unit be activated when the property is present.

\section{Relationship between Minout and the Gating Mechanism for Conditional Computation}
\label{sec:app_relation}
The original conditional computation paper and follow-up works introduce binary gating neurons to turn on/off computing neurons conditioned on inputs~\cite{Bengio2013}. In practice, \textit{sigmoid} activation functions are usually used in place of the binarization for the ease of training. Assuming both the computing neuron and gating neuron use \textit{sigmoid} activation functions, it can be written as:

\begin{equation}
y=\underbrace{\sigma(x^TW_1+b_1)}_\text{Computing neuron}\odot{}\underbrace{\sigma(x^TW_2+b_2)}_\text{Gating neuron}
\label{eqn:sup:gating}
\end{equation}
Note that the computing neuron and gating neuron are indistinguishable and can be swapped. We can generalize this into $y=\prod_j{\sigma(x^TW_j+b_j)}$. We can see that $y=0$ if there exists $j$ so that $\sigma(x^TW_j+b_j)=0$. And $y>0$ only when $\sigma(x^TW_j+b_j)>0$ for $\forall{j}$.

For clipped Minout:

\begin{equation}
   y=\min_{j\in{[1,k]}}\max(x^T W_{j}+b_{j}, 0)
   \label{eqn:sup:clipped_minout}
\end{equation}
It has a similar behavior, $y>0$ only when \textit{all} of $\max(x^T W_{j}+b_{j}, 0)>0$.

In fact, Eqn. \ref{eqn:sup:gating} and \ref{eqn:sup:clipped_minout} can be seen as AND functions, whose output is non-zero only when all the neurons are activated. The AND function is activated only when all conditions are satisfied. This means that if the minout function is properly trained only very specific examples will fall in $\mathcal{S}_1$, making a small set for rehearsal.

\section{Number of Rehearsed Examples}
\label{sec:app_num_rehearse}
One can consider the maxout unit as a detector of some property of the input. Inputs that have this property will fall within the convex set $\mathcal{S}_1$, we call them positive examples. Inputs that do not have this property will fall outside of the convex set, and we call them negative examples.

For a single maxout unit, the number of the rehearsed examples depends on the size of $\mathcal{S}_1$ and $\mathcal{S}_2$. $\mathcal{S}_2$ can be small if most negative examples are clipped at more than one neuron. The size of $\mathcal{S}_1$ depends on how many historical examples activates this maxout unit. In this paper, since each maxout unit corresponds directly to one of the categories, the size of $\mathcal{S}_1$ is approximately one tenth (for ten categories in MNIST) of the total training examples.

We can have more maxout units in the hidden layers in the future with a deep version of this idea. And we can study the relationship between the number of rehearsed examples and the number of maxout units.

\section{Stochastic Gradient Descent Fails on MNIST-\textit{ol}}
\label{sec:app_sgd_fail}

A 2-layer multilayer perceptron ($784\rightarrow128\rightarrow10$) with ReLU activations and a softmax loss is constructed and trained with SGD on MNIST-ol. It is compared against online MNIST with i.i.d. assumption, i.e. SGD with $1$ as the mini-batch size.

\begin{figure}[h]
    \centering
    \includegraphics[width=0.8\textwidth]{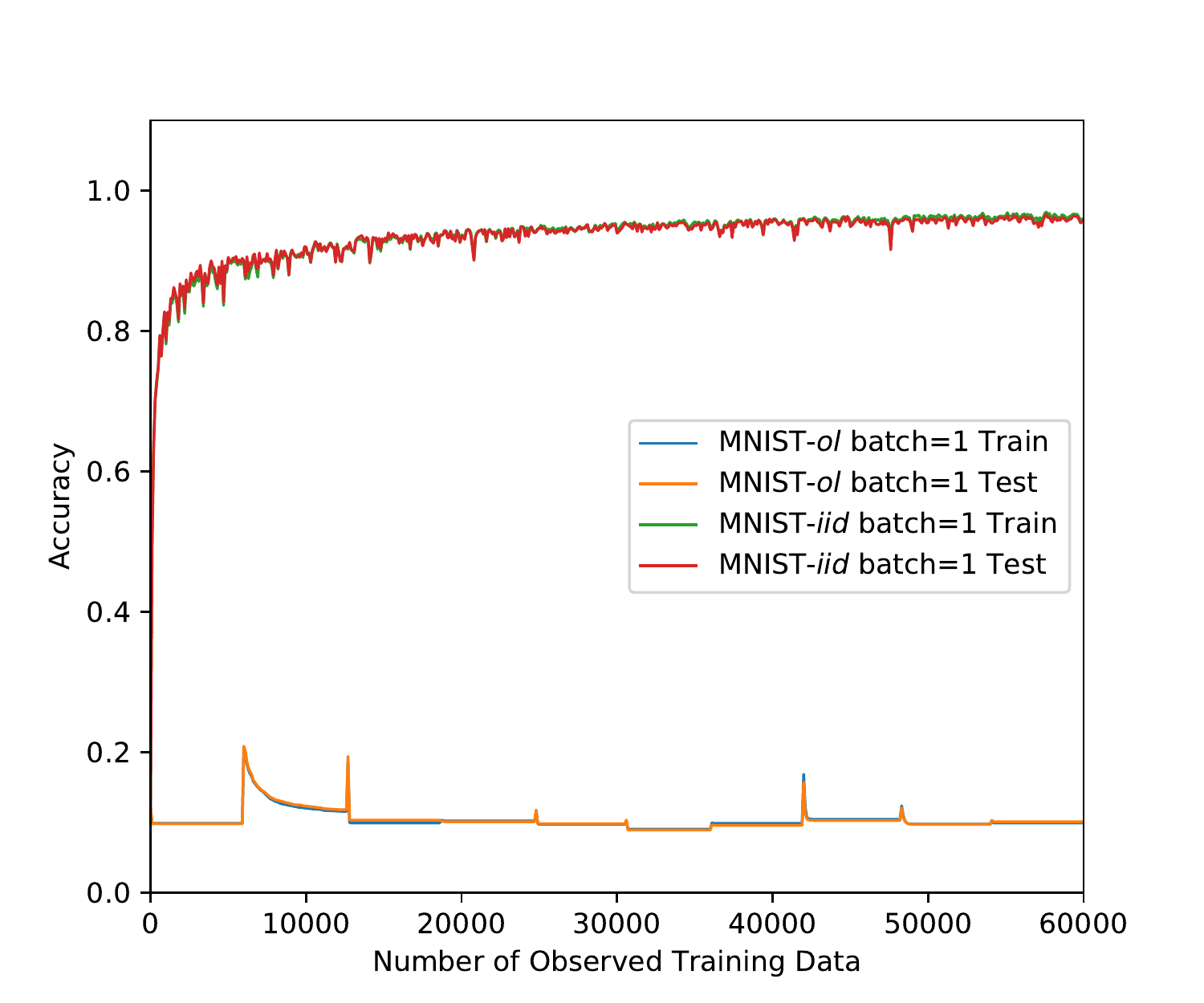}
    \caption{Training with Stochastic Gradient Descent fails on MNIST-ol.}
    \label{fig:mnist_ol_vs_iid}
\end{figure}

Fig. \ref{fig:mnist_ol_vs_iid} shows that SGD fails utterly on MNIST-ol where i.i.d. assumption is broken due to the ordering.
\end{appendices}

\end{document}